\documentclass[runningheads]{llncs}

\usepackage[T1]{fontenc}
\usepackage{graphicx}
\usepackage{multirow}
\usepackage{array}
\usepackage{booktabs} 
\usepackage{amsmath}
\usepackage{amssymb, bm}
\usepackage{makecell}
\usepackage[flushleft]{threeparttable}
\usepackage[title]{appendix}
\usepackage{tikz}
\usetikzlibrary{matrix}

\usepackage{amsmath}
\usepackage{pdfpages}
\usepackage{booktabs}
\usepackage{hyperref}
\DeclareMathOperator*{\argmax}{argmax} 

\begin{document}
\title{Terminating Differentiable Tree Experts}
\titlerunning{Terminating Differentiable Tree Experts}

\author{
    Jonathan Thomm\inst{1,2} \and
    Michael Hersche\inst{1}\and
    Giacomo Camposampiero\inst{1,2} \and
    Aleksandar Terzić\inst{1,2} \and
    Bernhard Schölkopf\inst{2,3} \and
    Abbas Rahimi\inst{1}
}
 \authorrunning{Thomm et al.}

\institute{
    IBM Research -- Zurich\and
    ETH Zürich \and
    Max Planck Institute for Intelligent Systems \\
    \email{abr@zurich.ibm.com}
}
\maketitle

\begin{abstract}
We advance the recently proposed neuro-symbolic Differentiable Tree Machine, which learns tree operations using a combination of transformers and Tensor Product Representations. We investigate the architecture and propose two key components. We first remove a series of different transformer layers that are used in every step by introducing a mixture of experts. This results in a Differentiable Tree Experts model with a constant number of parameters for any arbitrary number of steps in the computation, compared to the previous method in the Differentiable Tree Machine with a linear growth. Given this flexibility in the number of steps, we additionally propose a new termination algorithm to provide the model the power to choose how many steps to make automatically. The resulting Terminating Differentiable Tree Experts model sluggishly learns to predict the number of steps without an oracle. It can do so while maintaining the learning capabilities of the model, converging to the optimal amount of steps.

\end{abstract}
\section{Introduction}
\let\thefootnote\relax\footnotetext{\footnotesize\noindent 18th Conference on Neural-Symbolic Learning and Reasoning (NeSy 2024).}

Neuro-symbolic AI aims to combine the strengths of statistical AI, like machine learning, with the capabilities of symbolic AI, to address the weaknesses of each.
Recent neuro-symbolic AI methods exhibit notable benefits from a tight integration of low-level statistical perception and high-level reasoning, e.g., in various tasks demanding out-of-distribution (OOD) generalization~\cite{DeepProbLog_NEURIPS2018,ICML2018,NS_ConceptLearner_ICLR19,NeuralStack_NEURIPS2020,PrAE_CVPR21,hersche2023neuro,AAAI_2023}.
However, compared to pure neural approaches, neuro-symbolic AI suffers from the non-differentiability of inherently discrete symbolic operations, unlike real numerical values, which makes them incompatible with gradient-based learning methods. One solution can be reinforcement learning-based learning approaches; however, they suffer from ill-defined gradients~\cite{lorello2023nesylearning}. 
Another solution is to use a fully symbolic search. For instance, DreamCoder \cite{dreamcoder} builds an increasing library of functions from input-output examples and uses wake, abstraction, and sleep phases to find meaningful algorithm primitives for the library. Another example is the Neural-Symbolic Stack Machine~\cite{neural-symbolic-stack-machines} which uses a neural model giving instructions in a fixed language to be executed by a non-deep-learning part. This approach, however, is not differentiable, and therefore, during training, a correct execution trace is searched to obtain the multi-step training target. While such solutions excel at solving surprisingly complex examples, their base language might possess strong inductive biases, and it is unclear how to connect such a system to noisy or continuous signal streams.

One viable option is to integrate Tensor Product Representations (TPR)~\cite{Smolensky1990TensorPV} inside neural networks. 
TPR is a general schema for mapping symbolic structures to numerical vector representations that allow continuous manipulations.
Moreover, TPR can express a general formalization of compositional structure which is defined by an instance of a structure resulting from assigning a set of roles to particular fillers~\cite{NEWELL1980}: a role characterizes a position in the structure, and its filler is the substructure that occupies that position. 
In TPR, such compositional structure is constructed by binding the role vectors with the filler vectors using an outer product between the two vectors, which grows exponentially in dimension with the number of bound vectors.
To alleviate this explosion, there are closely related lossy compressed representational schemes~\cite{PlateHolographic1995,VSA_03} in which the vectors are closed under binding operations: i.e., all roles, fillers (substructures), and resulting compositional structures themselves can be represented by fixed-dimensional distributed vectors.

\sloppy 
The TPR has been integrated into various deep neural network architectures~\cite{Palangi_AAAI2018,schlag2019enhancing,chen2020mapping,jiang2021-enrichingTPR} (see~\cite{Neurocompositional} for an overview). Recently, TPR has been combined with a transformer to construct a Differentiable Tree Machine (DTM)~\cite{dtm} which learns sequences of operations on trees. The DTM represents trees using TPR tensors, defines a set of operations analogous to a subset of the Lisp~\cite{lisp-book} language on the TPR tree representations, and learns to execute the right Lisp operations to transform trees within multiple steps. A transformer model predicts the operations and arguments, and a TPR interpreter executes them in a differentiable manner. The DTM has shown OOD generalization in several tree transformation tasks, and provides a fairly general idea for learning of sequences of discrete symbolic operations on trees by making transformations and computations differentiable.

The DTM, on which our work is based on, faces several limitations. First, it uses a different parameterization of the transformer layer for each computation step. This means that as the number of steps of computation increases, the total model size increases linearly. Second, due to the first limitation, the DTM model only operates in a fixed number of steps. With that, the DTM also needs an oracle termination mechanism, i.e., one needs to know how many steps to compute for a given task. 
Further, the operations defined in the DTM have strong inductive biases towards certain tree operations. Therefore, the OOD performance shown in the original paper seems to exist because the model is invariant to those specific OOD cases. In contrast, its true OOD generalization is not as good in general as we demonstrate on a new tree reversal task in Section~\ref{sec:tree-reversal}.

In this work, we enhance the DTM architecture by making the following contributions that address these limitations:
\begin{itemize}
    \item We introduce a Mixture of Experts~\cite{switch-transformers,mistral-of-experts,sparse-moe} to the DTM architecture, resulting in a Differentiable Tree Experts (DTE) model. DTE uses the same parameters at every step and soft-chooses different weight combinations from a weight pool. This allows for more general use-cases as the DTE can iterate arbitrarily long without needing more parameters.
    
    \item We experiment with halting and introduce a new halting mechanism that adapts delicately enough to work with the slightly unstable training convergence of the DTM. This avoids the need for hyperparameter tuning and theoretically allows the model to learn how much computation is needed for a given task, similar to what has been used in Universal Transformers~\cite{universal-transformers}.
    \item Our DTE and Terminating DTE architectures exhibit similar ID and OOD performances as the DTM. However, the number of parameters of the proposed architectures scales constantly with the number of transformations required to solve the problem, compared to the linear scaling that characterizes the DTM.
    For example in Table~\ref{tab:rev-long-results}, DTM has $47\,M$ parameters compared to $27\,M$ for our DTE, with the difference getting larger with the number of steps. We also observe that sparsifying expert selection, which reduces the computational burden during training and inference, has no significant impact on performance.  
    \item We ablate the DTM and DTE architectures on a novel tree reversal task. We demonstrate that the OOD performance of the DTM architecture is attributable to invariance rather than generalization (see Section~\ref{sec:tree-reversal}).
\end{itemize}

\section{Background}
In this section, we briefly introduce Tensor Product Representations (TPRs) and the architecture of DTM.

\subsection{Tensor Product Representations}
Tensor Product Representation (TPR) provides a general
encoding of structured symbolic objects in vector space. A TPR consists of roles and fillers~\cite{Smolensky1990TensorPV}. While fillers describe the data, the roles define its context, and therefore, the TPR allows for a compositional symbolic representation via distributed vectors and tensors. 

To represent a symbolic object, one computes the outer product ($\otimes$) of the filler ($\mathbf{f}$) and the role ($\mathbf{r}$) vectors, resulting in a matrix $\mathbf{M}=\mathbf{f}\otimes \mathbf{r} = \mathbf{f} \mathbf{r}^T$. 
A set of $N$ symbolic objects is represented by the superposition of the role-filler products: 
\begin{align}\label{eq:role-filler-superpos}
    \mathbf{T} = \sum_{n=1}^{N} \mathbf{f}_n \otimes \mathbf{r}_n. 
\end{align}
If all role vectors are orthonormal, the individual fillers can be retrieved from $\mathbf{T}$ using an associative recall, e.g., 
\begin{align*}
    \mathbf{f}_i = \mathbf{T} \mathbf{r}_i = \sum_{n=1}^{N} \mathbf{f}_n \mathbf{r}^T_n \mathbf{r}_i. 
\end{align*}
The orthonormality constraint on the roles requires their dimensionality to be $\geq N$. 

Equation~\eqref{eq:role-filler-superpos} allows us to encode tree structures as well~\cite{dtm}. 
Let us consider an example of a binary tree of depth 4, illustrated in Fig.~\ref{fig:dtm-architecture}.
The leaves of the tree can be represented with TPR by $\mathbf{T} =  \mathbf{f}_{\text{some}} \otimes \mathbf{r}_{00}+ \mathbf{f}_{\text{sad}} \otimes \mathbf{r}_{100}  +  \mathbf{f}_{\text{sheep}} \otimes \mathbf{r}_{1100}$. 
Here, the subscript ($x$) of a role ($\mathbf{r}_x$) describes the path from the root to the leaf, e.g., $x=100$ describes the sequence right$\rightarrow$left$\rightarrow$left.

\subsection{Differentiable Tree Machine}
\begin{figure}[t]
    \centering
    \includegraphics[width=\textwidth]{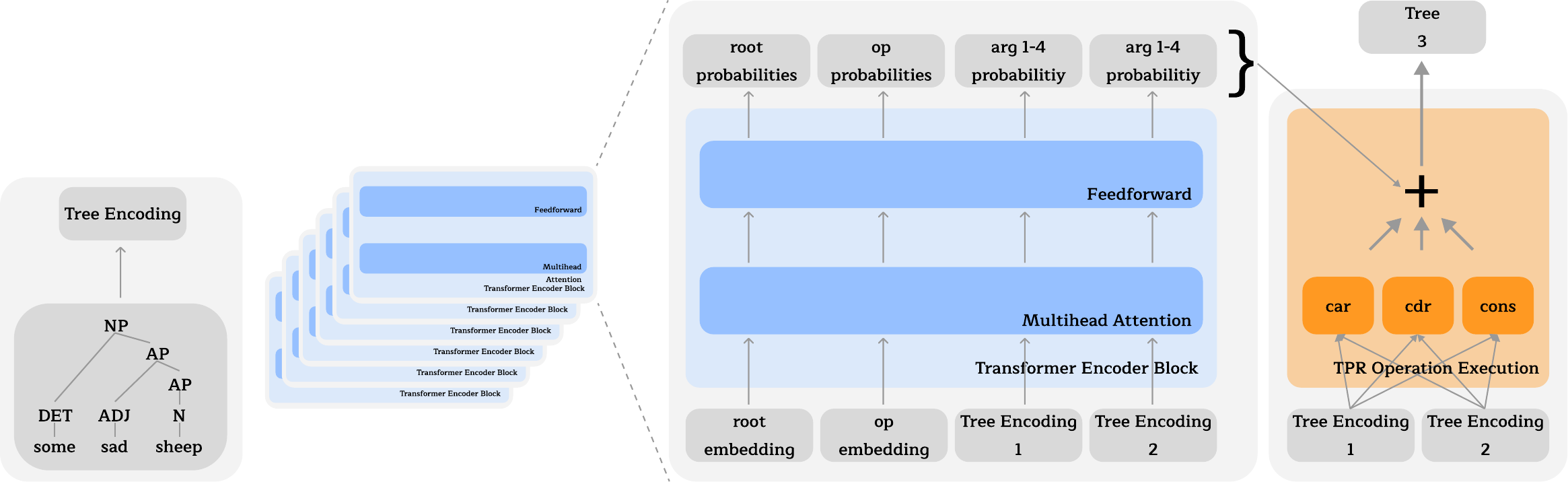}
    \caption{The DTM architecture. In each step, a new tree superposition is generated (in TPR) using a different transformer encoder layer for each step. The instruction probabilities are predicted by the transformer encoder layer. \textsc{car}, \textsc{cdr}, and \textsc{cons} are the three Lisp operations.}
    \label{fig:dtm-architecture}
\end{figure}

The Differentiable Tree Machine (DTM)~\cite{dtm} manipulates TPR-based tree representations using three Lisp operations: \textsc{car}, \textsc{cdr}, and \textsc{cons}. 
Given a tree ($\mathbf{T}$), Lisp \textsc{car} extracts the subtree that is the left child of the root by $\textsc{car}(\mathbf{T})=\mathbf{D}_0 \mathbf{T}$.
Here, $\mathbf{D}_0=\mathbf{I}\otimes \sum_x \mathbf{r}_x \mathbf{r}_{0x}^T$ is a linear operator that shifts all roles from the left subtree up to the root by one level, and $\mathbf{I}$ corresponds to the identity matrix on the filler space. 
Applying Lisp \textsc{car} to the example tree in Fig.~\ref{fig:dtm-architecture} would yield $\mathbf{T}_0=\textsc{car}(\mathbf{T})= \mathbf{f}_{\text{some}} \otimes \mathbf{r}_{0}$. 
Lisp \textsc{cdr} extracts the right child by $\textsc{cdr}(\mathbf{T})=\mathbf{D}_1 \mathbf{T}$, where $\mathbf{D}_1=\mathbf{I}\otimes \sum_x \mathbf{r}_x\mathbf{r}_{1x}^T$ is the linear operator that shifts all roles from the right subtree up to the root.
Finally, the \textsc{cons} operation takes two trees ($\mathbf{T}_0$ and $\mathbf{T}_1$) as arguments plus a new root node ($\mathbf{s}$) and assembles a new tree by $\textsc{cons}(\mathbf{T}_0, \mathbf{T}_1, 
\mathbf{s})= \mathbf{E}_0 \mathbf{T}_0+\mathbf{E}_1 \mathbf{T}_1+\mathbf{s}\otimes \mathbf{r}_{root}$. 
The linear operators $\mathbf{E}_0=\mathbf{I}\otimes \sum_x \mathbf{r}_{0x}\mathbf{r}_{x}^T$ and $\mathbf{E}_1=\mathbf{I}\otimes \sum_x \mathbf{r}_{1x}\mathbf{r}_{x}^T$ shift all roles to the left and right subtrees down to the leaves, respectively.

DTM generates a sequence of trees ($\mathbf{T}^{(0)}, \mathbf{T}^{(1)},...,\mathbf{T}^{(L)} $), where the initial tree ($\mathbf{T}^{(0)}$) is the source tree, and the final tree ($\mathbf{T}^{(L)}$) is the target tree (i.e., the result of the task).
DTM computes the tree at step $t$ as a convex combination of the results provided by the three Lisp operations, which creates a TPR representation of a new tree superposition: 
\begin{align*}
    \mathbf{T}^{(t+1)} = {w}^{(t)}_{\textsc{car}}\textsc{car}(\mathbf{T}^{(t)}_\textsc{car})+{w}^{(t)}_{\textsc{cdr}}\textsc{cdr}(\mathbf{T}^{(t)}_\textsc{cdr})+{w}^{(t)}_{\textsc{cons}}\textsc{cons}(\mathbf{T}^{(t)}_\textsc{cons,0},\mathbf{T}^{(t)}_\textsc{cons,1}, \mathbf{s}^{(t)}). 
\end{align*}
A transformer encoder layer (with the standard quadratic attention) predicts the weights (${w}^{(t)}_{\textsc{car}}, {w}^{(t)}_{\textsc{cdr}}, {w}^{(t)}_{\textsc{cons}}$) for the three Lisp operations, and their arguments ($\mathbf{T}^{(t)}_\textsc{car},\mathbf{T}^{(t)}_\textsc{cdr},\mathbf{T}^{(t)}_\textsc{cons,0}, \mathbf{T}^{(t)}_\textsc{cons,1}, \mathbf{s}^{(t)} $).
Given a list of previously generated trees plus the input tree, each tree is encoded as one token by encoding the tree TPR representation to a dense vector using a deep learning encoder~\cite{dtm-beta}.
Each tree argument for the next Lisp operation is computed as a weighted sum of all past trees, e.g., 
\begin{align*}
    \mathbf{T}_{\textsc{car}}^{(t)} = \sum_{i=0}^{t-1} a^{(i)}_{\textsc{car}}\mathbf{T}^{(i)}
\end{align*}

The weight of a tree ($a^{(i)}_{\textsc{car}}$) is predicted by the transformer encoder layer at the position where the corresponding tree ($\mathbf{T}^{(i)}$) was put into the transformer encoder layer. 
The operation weights and the new root weights for the \textsc{cons} operation are predicted at two special classification tokens given as input to the transformer encoder layer.

The DTM runs for a predefined number of steps (between 12 and 28, configured as a hyperparameter), each time generating a new tree superposition. Each step however uses a different transformer encoder layer to predict the next action. The output of the last step is taken as the model's answer. See Fig.~\ref{fig:dtm-architecture} for an architecture diagram. In the experiments reported in this paper, we use a newer version of the original DTM paper's code which has improved training stability and provided a lower parameter count~\cite{dtm-beta}.

\section{Terminating Differentiable Tree Experts}

\subsection{Differentiable Tree Experts}

This section presents the main contribution of our paper: Differentiable Tree Experts (DTE). Instead of learning a different transformer encoder in each step for DTM, one could share the weights. However, according to our experiments, using the same transformer encoder layer leads to a non-converging DTM. We, therefore, propose integrating a Mixture of Experts in the DTM architecture, which enables convergence again despite the weight sharing between each step. This means that in every step, the same router in DTE weights several experts (in our experiments, 16 experts) that then give proposals for the operation and the arguments. Those predictions are weighted, and then the DTE execution takes place. 

\begin{figure}[t]
    \centering
    \includegraphics[width=\textwidth]{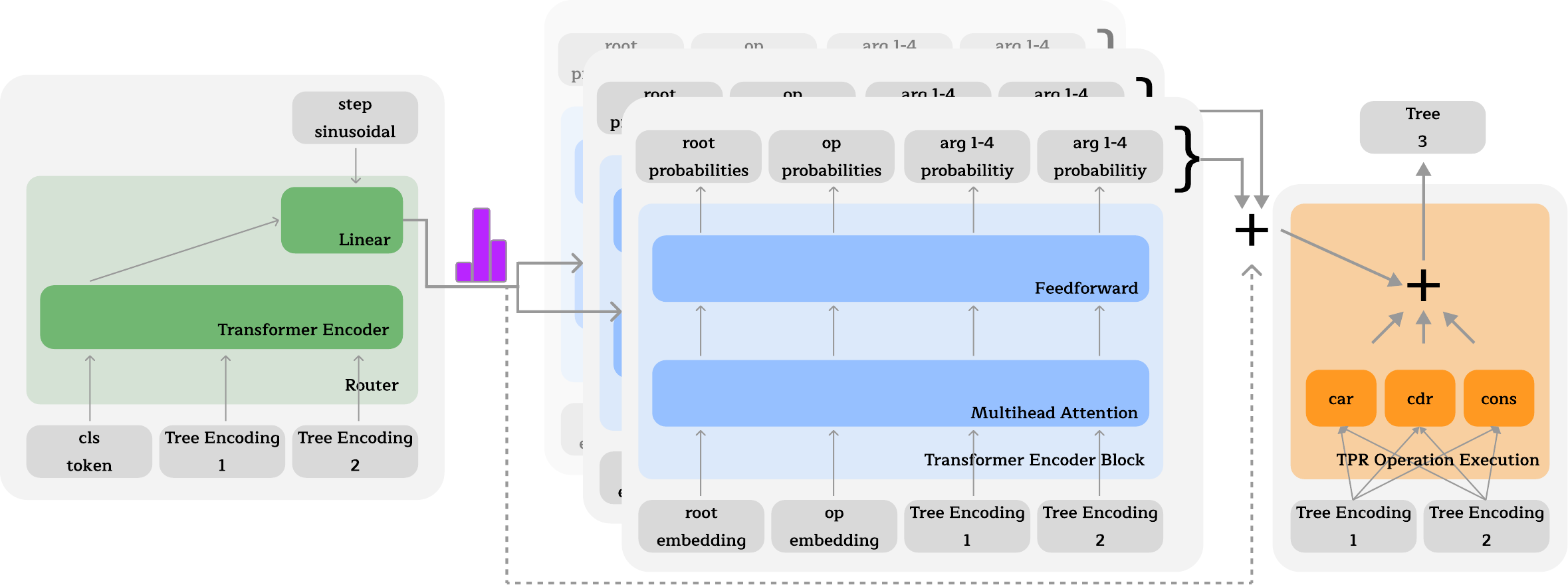}
    \caption{The architecture of the DTE. In each step, a new tree is generated using the same model. Our transformer encoder layer is now a Mixture of Experts (MoE) with the router itself being a combination of a transformer encoder layer and a linear map. The router chooses the expert weights, which then are used to weigh the outputs of each expert. In our sparse MoE ablations, only the top 4 experts are activated.}
    \label{fig:dte-architecture}
\end{figure}

In our router, a transformer encoder layer encodes all current trees. The current step is encoded as a sinusoidal positional encoding~\cite{attention-is-all-you-need}. From the concatenation of the tree encoding and the step encoding, the router probabilities are computed with a linear map. See Fig.~\ref{fig:dte-architecture} for an architecture diagram of DTE.

This architecture modification scales much better when the number of steps increases. While DTM needs an additional transformer encoder layer for every step, DTE stays constant in size. Together with a termination algorithm, this also allows for deciding the number of steps flexibly, e.g., allowing for a different number of steps during inference (in our experiments between 12 and 28).

\subsection{Sluggish Termination}

Several termination heuristics have been proposed in the literature \cite{adaptive-compute-time,pondernet}. For this work, we found termination inspired by speculative execution to work best for our Terminating DTE (TDTE). In general, the training convergence of DTE is brittle, as also observed with the DTM. In particular, changing the termination decision too often caused the model to not converge anymore. We, therefore, use two termination predictors. One predictor follows the other as soon as it is confident. This way, changes in the termination are only made if a certain confidence is reached. 

Let us denote $i_{\text{damp}}:=\text{argmax}_s\left(p_{\text{damp}}(s)\right)$, $i_{\text{expl}}:=\text{argmax}_s\left(p_{\text{expl}}(s)\right)$ the predictions of the two predictors, and $p(i_{\text{damp}}), p(i_{\text{expl}})$ the probabilities of the predictors at those indices. The probabilities over all steps sum up to 1 for each predictor.
We define the loss label (i.e., the target step) of the two termination predictors as:

\newpage

\begin{figure}[t]
    \centering
    \includegraphics[width=\textwidth]{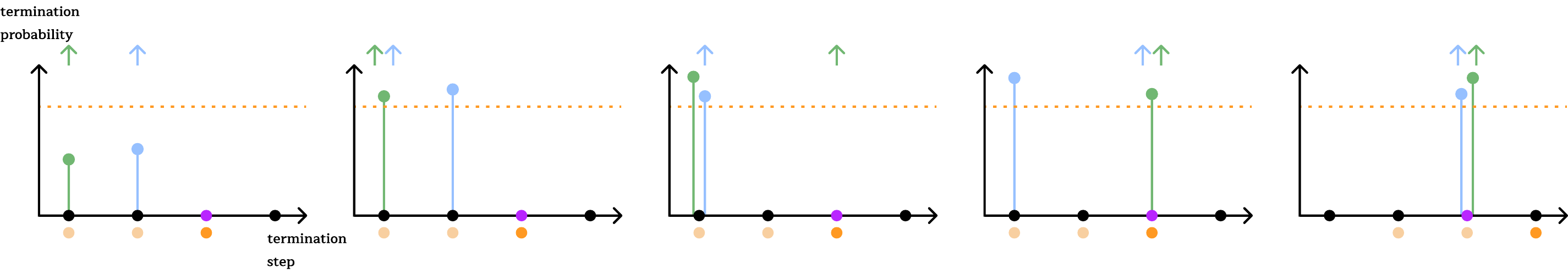}
    \caption{Cases of the sluggish termination losses. The two arrows indicate the labels of the termination predictors in each case. The cases are given by whether the predictors are below or above the yellow confidence threshold. The orange dots show where the main model loss is and the two (relatively small) residual losses. The purple dot is the best local termination (e.g., best loss with some step penalty). The green predictor is called the ``explorer'', the blue one the ``damper'', as it will start to follow the explorer when the explorer becomes confident and otherwise stays where it is.}
    \label{fig:sluggish-termination}
\end{figure}

\begin{align}
    \label{eq:predictors1}
    y_{\text{damp}} &= \begin{cases}
        i_{\text{expl}} & p(i_{\text{expl}})\geq 0.8 \\
        i_{\text{damp}} & \text{otherwise}
    \end{cases} \\
    \label{eq:predictors2}
        y_{\text{expl}} &= \begin{cases}
        \underset{s\in S}{\argmax}\left(\text{loss}(s)*0.9^{\text{idx}(s)}\right) & \text{if } p(i_{\text{damp}})\geq 0.8 \land i_{\text{damp}}=i_{\text{expl}} \\
        i_{\text{expl}} & \text{otherwise}
    \end{cases}
\end{align}

As shown in Equations~\eqref{eq:predictors1} and \eqref{eq:predictors2}, we use a confidence threshold of $0.8$ which was determined by a grid search based on the training convergence. $S$ denotes the local set of choices around the current prediction, i.e. $S=\{i_{\text{damp}}-4,i_{\text{damp}},i_{\text{damp}}+5\}$ and $\text{idx}(i_{\text{damp}}-4)=0$, $\text{idx}(i_{\text{damp}})=1$, $\text{idx}(i_{\text{damp}}+5)=2$. The choices in $S$ are hyperparameters that worked well in practice.

We learn one termination for the DTM on each task. Each predictor consists of a series of constant parameters, one for each potential step, which is enough for the datasets investigated here and in the datasets used in~\cite{dtm}. By scaling the termination parameters (and initializing them to small values) by a large factor, one can make sure that the gradient updates are high enough for those single values. The method can also be applied to sample-wise predictors from the main model, which remains to be explored in future work.

Fig.~\ref{fig:sluggish-termination} visualizes Equations~\eqref{eq:predictors1} and \eqref{eq:predictors2}, i.e. which losses are applied to the two termination predictors in which cases. The exploration predictor (blue) will stay where it is until the damping predictor (green) is confident. When this happens, the exploration predictor stays where it is, and the damping predictor follows. As soon as both are at the same place and confident, the exploration predictor starts exploring for better termination options again.

To compute which of the three considered termination steps is the best (see Equation~\ref{eq:predictors2} top case), we calculate the loss at each of the three steps and deduct a small multiplicative factor for later termination.
This way the model will choose to terminate earlier if iterating longer does not bring significant improvements. We use cross-entropy loss to train the predictors, which are single numbers for each step and predictor. Our main loss is on the last choice, i.e., after the termination (see Fig.~\ref{fig:sluggish-termination}), to make sure the model can learn to do intermediate work at the current termination and then decide to terminate later.

The DTM and DTE architectures use the last tree as the final model output. When the termination changes, a step previously computing the final answer now computes an intermediate result. This could lower training performance; therefore, having a separate read-out transformer layer and choosing the termination over intermediate results only, could be an improvement for future work.

\section{Experiments}\label{sec:main-results}
\begin{figure}[t]
    \centering
    \includegraphics[width=\textwidth]{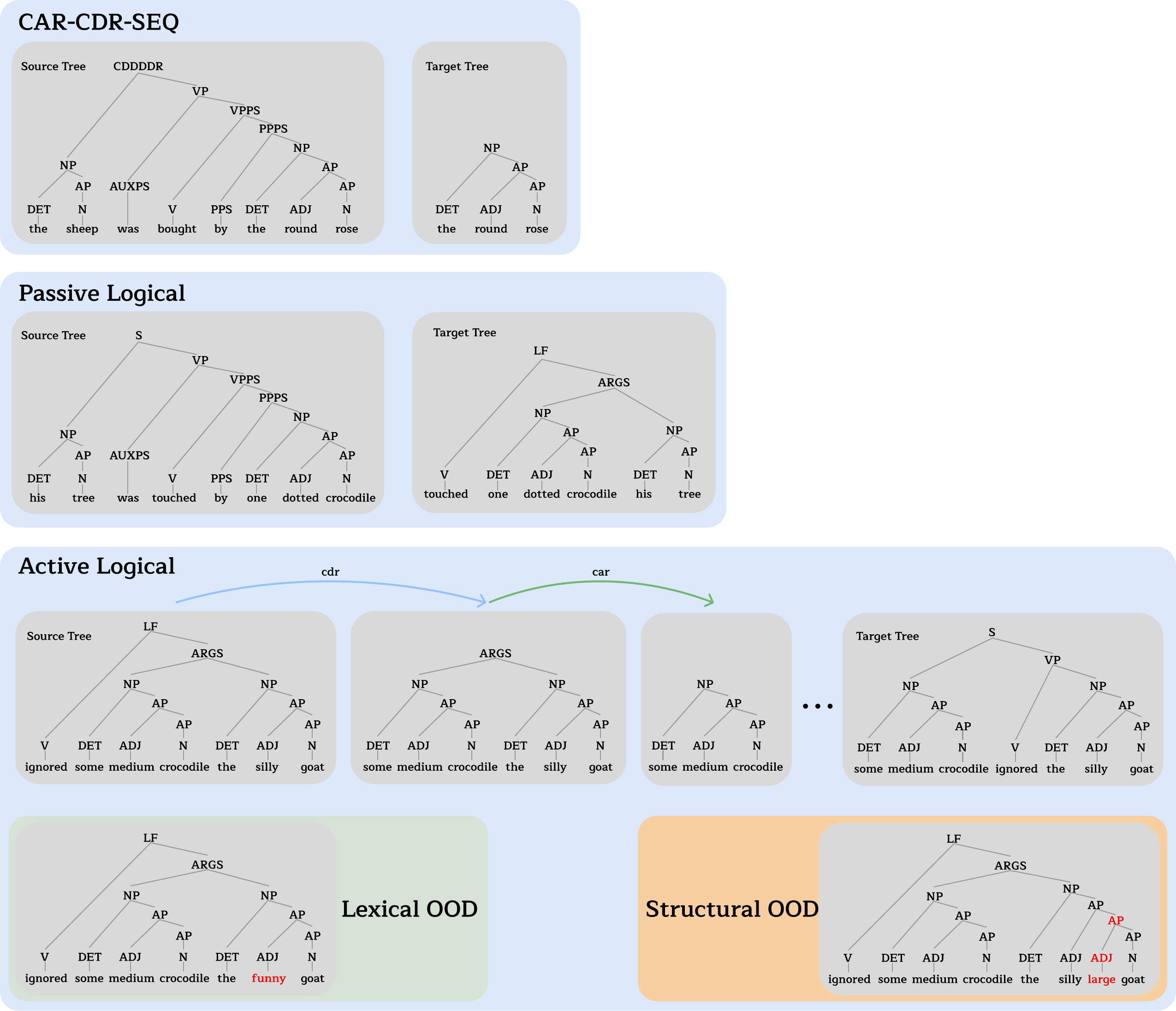}
    \caption{
    Examples of the \textsc{Car-Cdr-Seq}, \textsc{Passive}$\leftrightarrow$\textsc{Logical}, and \textsc{Active}$\leftrightarrow$\textsc{Logical} dataset~\cite{dtm}. The model has to transform a source tree to the target tree. 
    For the \textsc{Passive}$\leftrightarrow$\textsc{Logical} case we show the intermediate trees that the model could produce to get to the target tree. Moreover, we show an example of lexical generalization that uses unseen adjectives (in this case ``funny''), as well as one for the structural generalization test set that adds additional adjectives.}
    \label{fig:active-logical-visualization}
\end{figure}

\begin{table}[ht]
    \centering
    \setlength{\tabcolsep}{9pt}
    \renewcommand{\arraystretch}{1.1}
    \begin{tabular}{lccc}
        \toprule
        Dataset & DTM & DTE & TDTE  \\
        \midrule
        \textbf{\textsc{Car-Cdr-Seq}} & && \\
        -\textsc{train} & 0.99 ± 0.00 & 1.00 ± 0.00 & 0.99 ± 0.00 \\
        -\textsc{test ID} & 1.00 ± 0.00 & 1.00 ± 0.00 & 1.00 ± 0.00 \\
        -\textsc{test OOD lexical} & 0.99 ± 0.01 & 0.99 ± 0.01 & 0.98 ± 0.01 \\
        -\textsc{test OOD structural} & 0.96 ± 0.01 & 0.97 ± 0.01 & 0.94 ± 0.03 \\
        \textbf{\textsc{Active}$\leftrightarrow$\textsc{Logical}} & && \\
        -\textsc{train} & 1.00 ± 0.00 & 1.00 ± 0.00 & 1.00 ± 0.00 \\
        -\textsc{test ID} & 1.00 ± 0.00 & 1.00 ± 0.00 & 1.00 ± 0.00 \\
        -\textsc{test OOD lexical} & 1.00 ± 0.00 & 1.00 ± 0.00 & 0.94 ± 0.06 \\
        -\textsc{test OOD structural} & 1.00 ± 0.00 & 1.00 ± 0.00 & 0.96 ± 0.05 \\
        \textbf{\textsc{Passive}$\leftrightarrow$\textsc{Logical}} & && \\
        -\textsc{train} & 1.00 ± 0.00 & 1.00 ± 0.00 & 1.00 ± 0.00 \\
        -\textsc{test ID} & 1.00 ± 0.00 & 1.00 ± 0.00 & 1.00 ± 0.00 \\
        -\textsc{test OOD lexical} & 1.00 ± 0.00 & 1.00 ± 0.00 & 1.00 ± 0.00 \\
        -\textsc{test OOD structural} & 1.00 ± 0.00 & 1.00 ± 0.00 & 0.99 ± 0.02 \\
        \textbf{\textsc{Active}\&\textsc{Passive}$\rightarrow$\textsc{Logical}} & && \\
        -\textsc{train} & 1.00 ± 0.00 & 1.00 ± 0.00 & 1.00 ± 0.00 \\
        -\textsc{test ID} & 1.00 ± 0.00 & 1.00 ± 0.00 & 1.00 ± 0.00 \\
        -\textsc{test OOD lexical} & 1.00 ± 0.00 & 1.00 ± 0.00 & 1.00 ± 0.00 \\
        -\textsc{test OOD structural} & 0.98 ± 0.02 & 0.99 ± 0.03 & 0.98 ± 0.02 \\
        \bottomrule
        \vspace{0.0cm}
    \end{tabular}
    \caption{Performance of the reproduced DTM (our runs) and our new architectures DTE and Terminating DTE (TDTE) on the same set of tasks used for evaluating DTM~\cite{dtm}.}
    \label{tab:main-results}
    \vspace{-0.5cm}
\end{table}

We evaluate our Differentiable Tree Experts (DTE) and the Terminating DTE (TDTE) on the same set of four tasks used for the evaluation of DTM~\cite{dtm}. 
Fig.~\ref{fig:active-logical-visualization} visualizes examples of these tasks.
The first task, \textsc{Car-Cdr-Seq}, encodes left-subtree and right-subtree Lisp operations in the root node of the input tree. Those should be executed, and the resulting sub-tree is the answer. The task \textsc{Active}$\leftrightarrow$\textsc{Logical} task contains sentence grammar trees in either active or logical form; the task is to transform the tree into the other grammatical form. The \textsc{Passive}$\leftrightarrow$\textsc{Logical} task is analogous, having a sentence grammar in passive form instead of active. 
Finally, the \textsc{Active}\&\textsc{Passive}$\rightarrow$\textsc{Logical} task contains either active or passive sentence grammar trees and the target is the logical form. All tasks come with an ID test set and two OOD test sets. The lexical OOD set contains trees with adjectives never seen on the leaves. The structural OOD test set contains trees where additional adjectives are added.

While DTM and DTE use the same hidden dimension in the transformer encoder layers (64), we observe that TDTE requires a larger one (256) in order to obtain performance competitive with DTM and DTE. For the Mixture of Experts router, unlike other works~\cite{switch-transformers}, we do not use a load balancing loss for the router as it did not improve performance. In practice, the DTE and TDTE have more operations during training because of the (parallelizable) mixture of experts and the additional router. In more complex settings where the termination is different for different samples, the DTE and TDTE can become more efficient. The mixture of experts models become smaller in the number of parameters than the DTE for high depth.

We used 16 experts for all of our experiments which turned out to work well. This results in 17 transformer layer weights (one additional for the router) compared to 16-28 layers of the DTM for the datasets in Table \ref{tab:main-results} and up to 56 in Section \ref{sec:tree-reversal}.

Table~\ref{tab:main-results} shows the comparison results. 
We use the DTM as the strongest baseline, as it outperformed vanilla Transformers, Tree Transformers, LSTMs, and Tree2Tree LSTMs~\cite{dtm}.
Similarly to the original DTM paper, runs that reached validation accuracy less than $90\%$ were excluded because the training is unstable for all models. 5 remaining runs were taken per data point. As shown, both DTE and TDTE perform very similarly compared to DTM. 
At the same time, our DTE and TDTE require a constant number of parameters with respect to the number of steps, whereas DTM's model size grows linearly with the number of steps.
Moreover, TDTE does not access the oracle knowledge of the required number of steps, yet it performs on par with DTM.

\subsection{Ablation: Sparse Mixture of Experts}

\begin{table}[t]
    \centering
    \setlength{\tabcolsep}{9pt}
    \renewcommand{\arraystretch}{1.1}
    \begin{tabular}{lccc}
        \toprule
        Dataset & DTE & Sparse DTE & Sparse TDTE  \\
        \midrule
        \textbf{\textsc{Car-Cdr-Seq}} & && \\
        -\textsc{train} & 1.00 ± 0.00 & 0.99 ± 0.01 & 0.99 ± 0.01 \\
        -\textsc{test ID} & 1.00 ± 0.00 & 1.00 ± 0.01 & 1.00 ± 0.01 \\
        -\textsc{test OOD lexical} & 0.99 ± 0.01 & 0.99 ± 0.01 & 0.99 ± 0.01 \\
        -\textsc{test OOD structural} & 0.97 ± 0.01 & 0.96 ± 0.03 & 0.93 ± 0.03 \\
        \textbf{\textsc{Active}$\leftrightarrow$\textsc{Logical}} & && \\
        -\textsc{train} & 1.00 ± 0.00 & 1.00 ± 0.00 & 1.00 ± 0.00 \\
        -\textsc{test ID} & 1.00 ± 0.00 & 1.00 ± 0.00 & 1.00 ± 0.00 \\
        -\textsc{test OOD lexical} & 1.00 ± 0.00 & 0.94 ± 0.14 & 0.98 ± 0.05 \\
        -\textsc{test OOD structural} & 1.00 ± 0.00 & 0.98 ± 0.04 & 0.98 ± 0.04 \\
        \textbf{\textsc{Passive}$\leftrightarrow$\textsc{Logical}} & && \\
        -\textsc{train} & 1.00 ± 0.00 & 1.00 ± 0.00 & 1.00 ± 0.00 \\
        -\textsc{test ID} & 1.00 ± 0.00 & 1.00 ± 0.00 & 1.00 ± 0.00 \\
        -\textsc{test OOD lexical} & 1.00 ± 0.00 & 1.00 ± 0.00 & 0.90 ± 0.10 \\
        -\textsc{test OOD structural} & 1.00 ± 0.00 & 0.95 ± 0.09 & 0.99 ± 0.02 \\
        \textbf{\textsc{Active}\&\textsc{Passive}$\rightarrow$\textsc{Logical}} & && \\
        -\textsc{train} & 1.00 ± 0.00 & 1.00 ± 0.00 & 1.00 ± 0.00 \\
        -\textsc{test ID} & 1.00 ± 0.00 & 1.00 ± 0.00 & 1.00 ± 0.00 \\
        -\textsc{test OOD lexical} & 1.00 ± 0.00 & 0.99 ± 0.01 & 0.96 ± 0.08 \\
        -\textsc{test OOD structural} & 0.99 ± 0.03 & 0.97 ± 0.07 & 0.97 ± 0.06 \\
        \bottomrule
        \vspace{0.0cm}
    \end{tabular}
    \caption{Performance of the DTE and Terminating DTE with sparse Mixture of Experts (with 4 out of 16 experts being active each time). Except for outliers, the performance of the sparse models is very similar and within the same range as the dense model.}
    \vspace{-0.5cm}
    \label{tab:sparse-moe}
\end{table}

One can further reduce the computational amount required for the DTE during training and inference by introducing sparsity in the selection of experts. To this end, we always select only the top four experts and normalize the corresponding selection weights using the softmax function.

Table~\ref{tab:sparse-moe} shows the results. The performance of DTE is the same as in Table~\ref{tab:main-results} and only repeated for the reader's convenience. Again, we observe very similar performance, with small OOD performance reductions for the Sparse TDTE and an outlier in Test OOD Lexical for the Sparse DTE. This shows that sparse experts, in principle, also work with this model and can provide up to four times faster training and inference speed in the deep learning part.

\subsection{New Task: Tree Reversal}

\label{sec:tree-reversal}

\begin{figure}[t]
    \centering
    \includegraphics[width=\textwidth]{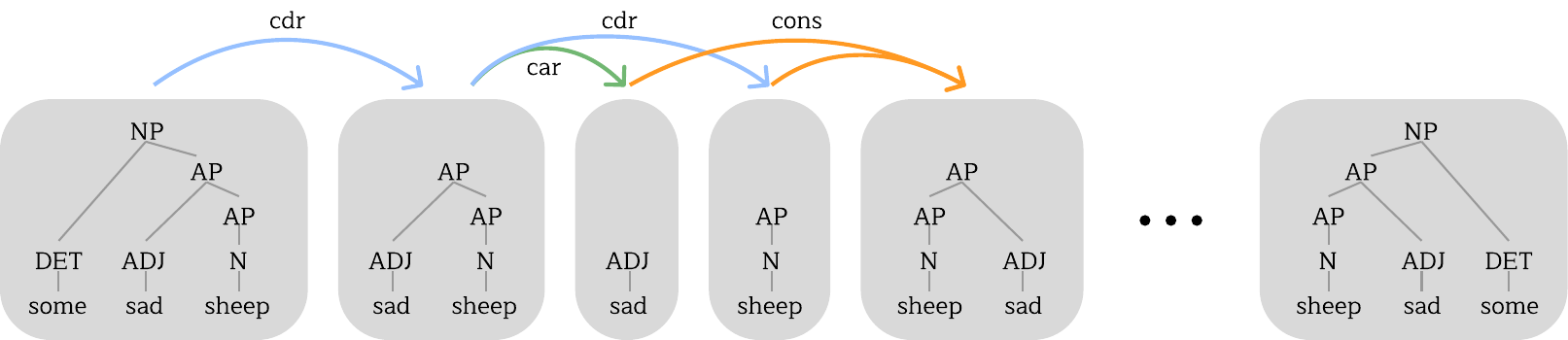}
    \caption{Visualization of how the DTM and our (T)DTE can solve our novel tree reversal task. As shown, with the three operations \textsc{cdr}, \textsc{car}, and \textsc{cons}, reversing a tree requires several steps and more with growing tree size since every child of a branching node needs to be extracted to assemble the tree in reverse order afterward.}
    \label{fig:tree-reversal}
\end{figure}

\begin{table}[t]
    \centering
    \setlength{\tabcolsep}{5pt}
    \renewcommand{\arraystretch}{1.2}
    \begin{tabular}{lccccc}
        \toprule
        Dataset & DTM & DTE & TDTE & SDTE & STDTE \\
        \midrule
        \textbf{\textsc{Reverse}} & && &&\\
        -\textsc{train} & 0.25±0.18 & 0.37±0.34 & 0.02±0.04 & \textbf{0.80}±0.19 & 0.16±0.23\\
        -\textsc{test ID} & 0.23±0.17 & 0.37±0.34 & 0.02±0.04 &\textbf{0.79}±0.20 & 0.14±0.22\\
        -\textsc{OOD lexical} & 0.02±0.03 & 0.05±0.06 & 0.00±0.00 & \textbf{0.39}±0.39 & 0.09±0.16\\
        -\textsc{OOD structural} & 0.00±0.00 & 0.00±0.00 & 0.00±0.00 & 0.00±0.00& 0.00±0.00\\
        \bottomrule
        \vspace{0.0cm}
    \end{tabular}
    \caption{Performance of the original DTM architecture and our DTE and TDTE architectures (both dense and sparse) on the tree reversal task. The training performance is highly dependent on the seed. Since these models are not invariant to this structural OOD, we observe 0\% performance everywhere on this split.}
    \vspace{-0.5cm}
    \label{tab:rev-results}
\end{table}

\begin{table}[t]
    \centering
    \setlength{\tabcolsep}{9pt}
    \renewcommand{\arraystretch}{1.1}
    \begin{tabular}{lccc}
        \toprule
        Dataset & DTM & DTE & TDTE  \\
        \midrule
        \textbf{\textsc{Reverse (56 steps)}} &&& \\
        -\textsc{train} &0.82&\textbf{1.00}& 0.00 \\
        -\textsc{test ID} &0.79&\textbf{1.00}&0.00 \\
        -\textsc{test OOD lexical} &0.05&\textbf{1.00}& 0.00\\
        -\textsc{test OOD structural} &0.00&0.00&0.00 \\
        \bottomrule
        \vspace{0.0cm}
    \end{tabular}
    \caption{DTM and DTE architectures on the reversal task when given many steps. The DTM and DTE architectures all use a hidden dimension size of 256 here (instead of 64), because it performed better. Only one run is taken per model.}
    \vspace{-0.5cm}
    \label{tab:rev-long-results}
\end{table}

This section evaluates DTM and TDTE on a new tree reversal task. The model gets a tree and has to reverse it exactly. This means that every inner node that has two children has to be extracted and reversed. Because the trees sometimes differ and subtrees to a higher depth have to be extracted, this task is more challenging. Especially the structural OOD now requires more and different operations. The input trees are the same as the input trees of the \textsc{Active}$\leftrightarrow$\textsc{Logical} task. See Fig. \ref{fig:tree-reversal} for a visualization with 28 steps.

As shown in Table \ref{tab:rev-results}, the models are able to learn tree reversals partially, which is a good sign, since the model needs to choose different Lisp operations for different samples. The structural OOD test set now needs other Lisp instructions and here we see that the model does not generalize to them at all.

We also train the model on many more steps (56 steps, see Table \ref{tab:rev-long-results}). This should give the model enough room to execute enough Lisp operations to reverse the whole tree.
The ID accuracy increases significantly, especially for DTE, which outperforms DTM while having a lower parameter count ($27\,M$ vs $47\,M$). The good ID performance confirms that the model can learn more complicated tree transformations, and its application to more complex datasets looks promising in general. The TDTE does not learn. 

Furthermore, it can be observed that the model generalizes to the lexical OOD set, but not structurally. This can be explained because the model can entirely ignore the precise adjectives, and as they are the tree leaves, they are unlikely to be extracted by the Lisp operations. In other words, the models are mostly invariant to OOD samples also for the reverse task. However, this is not the case for the structural OOD (which adds an additional adjective to the tree and therefore requires reversing two or more adjectives), and we see that once the invariance drops, there is no generalization at all anymore for all models in this experiment.

\section{Discussion}
DTM is a neuro-symbolic method for solving tree-to-tree transformation tasks, effectively combining a neural controller (i.e., a transformer) with a symbolic manipulator (TPR). 
The TPR engine performs symbolic manipulations in a continuous vector space, which allows a convex combination of different discrete operations (i.e., \textsc{Car}, \textsc{Cdr}, and \textsc{Cons}) and their operators (i.e., weighted superposition of past trees).
This yields a fully differentiable neuro-symbolic architecture that can be trained end-to-end, without requiring reinforcement learning techniques. 
We further enhanced DTM by introducing a mixture of experts and a novel automatic termination method, which reduces both the number of parameters and the required knowledge about the number of steps. 

Both the DTM and TDTE still face some inherent limitations, which we elaborate on in this section. 
Future work addressing the following limitations would notably enhance the DTM/TDTE approach.

\subsection{Limited Lisp operations}
DTM and TDTE models focus on binary tree-to-tree transformation tasks that can be solved with a limited subset of Lisp operations (\textsc{Car}, \textsc{Cdr}, and \textsc{Cons}). 
We have seen that the tasks presented in~\cite{dtm} can be solved with a relatively low number of sequential operations. 
By introducing the novel tree reversal task, however, we could show that DTE can indeed learn to execute longer sequences of the three operations. 

Distributed representations are not restricted to the three Lisp operations. 
For example, leveraging fractional power encoding (FPE)~\cite{PlateHolographic1995} would allow one to perform arithmetic operations. 
In fact, FPE has been applied in probabilistic abductive reasoning to solve Raven's progressive tests~\cite{hersche2023neuro}. 
Representing both logical and arithmetic rules with distributed representations yielded a differentiable and fast symbolic engine, which can even learn the underlying rules~\cite{hersche2023_learnVSA}. 
Using FPE to support arithmetic operations in DTM or TDTE would further enrich the architectures and is an interesting avenue for future work.

\subsection{OOD generalization}
Our novel tree reversal task reveals an important limitation of the DTM and TDTE models in OOD generalization, showing that when the data does not fit the strong inductive biases of Lisp operations, the models do not seem to generalize well.
The three Lisp operations have strong inductive biases to
allow certain tasks to be done in very few steps. However, other tasks, such as tree reversal, require many steps. The fixation on the Lisp operations, therefore, is a limitation for other tree-to-tree tasks than the tasks evaluated in the previous sections.

Further, the tree-to-tree tasks evaluated in the original paper~\cite{dtm} require mostly the same sequence of the three Lisp operations for all samples including the OOD test sets. This allows DTM to generalize very well---the model is invariant to the OOD variants tested, rather than generally having excellent OOD capabilities. For \textsc{Active}$\leftrightarrow$\textsc{Logical} and \textsc{Passive}$\leftrightarrow$\textsc{Logical}, the model only has to detect if the input is a sentence tree in active or in logical form, which is possible by simply looking at the children of the root node. These nodes are not really out of distribution in the OOD splits, where the only trees changed are the ones close to the leaves. See Fig.~\ref{fig:active-logical-visualization} for a visualized example.
In \textsc{Active}\&\textsc{Passive}$\rightarrow$\textsc{Logical} the same holds and the task \textsc{Car-CDR-Seq} is only about executing the steps encoded in the root node of the tree. Those operations encoded in the root are ID in the OOD split samples (the trees are different only close to the leaves but the model only needs to look at the root).

For example, in the lexical generalization test, new adjective fillers are used at the tree leaves. During execution, these values are never taken into account during execution, as the TPR engine is invariant to the filler values.
In the structural generalization test set, additional adjectives are added to the tree, while the operations that the model needs to learn are still the same, i.e., the transformer can ignore the bigger tree and only look at the inner nodes, which are still exactly equal.
Such invariance is also the case for the lexical generalization of the tree reversal task but not for the structural OOD. This is because for the tree reversal, the fillers of the leaves do not matter; the model does not need to reverse the leaves and only needs to reverse the tree based on the inner nodes. However, when the structure of the tree contains additional adjectives its shape changes, and with it the inner nodes. Therefore, more different operations are necessary to reverse the tree.

\subsection{Training stability}
Although the model combines deep learning and TPR successfully, the brittle training convergence is still a limitation. 
Removing this issue would be critical to allow the broader applicability of these hybrid models. 
The dependence on the initialization suggests that the optimization landscape is very non-convex or that the correct gradients are vanishing. Given the nature of the model, which linearly superposes all operations, the latter seems especially plausible. Investigating this problem and potentially finding improved optimizers or in-model solutions would make the applications of deep learning combined with TPR much more attractive to a broader audience.
To avoid vanishing gradients, one could also introduce a hybrid optimization including elements to limit the number of superpositions and, therefore, strengthen the remaining ones.

\section{Conclusion}

We have introduced Terminating Differentiable Tree Experts (TDTE) which enhances the recently proposed DTM architecture. Our improvements allow the model to scale constantly when the depth of computation increases. Based on this, we are further able to introduce a new halting mechanism that changes its decisions slowly and looks ahead multiple steps to be more precise and have less impact on the model performance. This method makes it possible to learn the right termination without having access to an oracle termination information within the training data (which is usually not given). 

\section*{Acknowledgment}
This work is supported by the Swiss National Science foundation (SNF), grant 200800. We thank Paul Soulos for providing the DTM code as well as inputs on the work.

\bibliographystyle{splncs04}

\begin{thebibliography}{10}
\providecommand{\url}[1]{\texttt{#1}}
\providecommand{\urlprefix}{URL }
\providecommand{\doi}[1]{https://doi.org/#1}

\bibitem{DeepProbLog_NEURIPS2018}
Manhaeve, R., Dumancic, S., Kimmig, A., Demeester, T., De~Raedt, L.: {DeepProbLog}: Neural probabilistic logic programming. In: Advances in Neural Information Processing Systems (NeurIPS). vol.~31 (2018)

\bibitem{ICML2018}
Xu, J., Zhang, Z., Friedman, T., Liang, Y., Van~den Broeck, G.: A semantic loss function for deep learning with symbolic knowledge. In: Proceedings of the 35th International Conference on Machine Learning (ICML). vol.~80, pp. 5502--5511 (2018)

\bibitem{NS_ConceptLearner_ICLR19}
Mao, J., Gan, C., Kohli, P., Tenenbaum, J.B., Wu, J.: The neuro-symbolic concept learner: Interpreting scenes, words, and sentences from natural supervision. In: International Conference on Learning Representations (ICLR) (2019)

\bibitem{NeuralStack_NEURIPS2020}
Chen, X., Liang, C., Yu, A.W., Song, D., Zhou, D.: Compositional generalization via neural-symbolic stack machines. In: Advances in Neural Information Processing Systems (NeurIPS). vol.~33, pp. 1690--1701 (2020)

\bibitem{PrAE_CVPR21}
Zhang, C., Jia, B., Zhu, S.C., Zhu, Y.: Abstract spatial-temporal reasoning via probabilistic abduction and execution. In: Proceedings of the IEEE Conference on Computer Vision and Pattern Recognition (CVPR) (2021)

\bibitem{hersche2023neuro}
Hersche, M., Zeqiri, M., Benini, L., Sebastian, A., Rahimi, A.: A neuro-vector-symbolic architecture for solving {R}aven’s progressive matrices. Nature Machine Intelligence  \textbf{5}(4),  363--375 (2023)

\bibitem{AAAI_2023}
Liu, A., Xu, H., Van~den Broeck, G., Liang, Y.: Out-of-distribution generalization by neural-symbolic joint training. In: Proceedings of the AAAI Conference on Artificial Intelligence. vol.~37, pp. 12252--12259 (2023)

\bibitem{lorello2023nesylearning}
Lorello, L.S., Lippi, M.: The challenge of learning symbolic representations. In: Proceedings of the 17th International Workshop on Neural-Symbolic Learning and Reasoning (NeSy) (2023)

\bibitem{dreamcoder}
Ellis, K., Wong, L., Nye, M., Sabl-Meyer, M., Cary, L., Pozo, L., Hewitt, L., Solar-Lezama, A., Tenenbaum, J.: {DreamCoder}: growing generalizable, interpretable knowledge with wake–sleep bayesian program learning. Philosophical Transactions of the Royal Society A: Mathematical, Physical and Engineering Sciences  \textbf{381} (06 2023)

\bibitem{neural-symbolic-stack-machines}
Chen, X., Liang, C., Yu, A.W., Song, D., Zhou, D.: Compositional generalization via neural-symbolic stack machines. In: Proceedings of the 34th International Conference on Neural Information Processing Systems (NeurIPS). Curran Associates Inc., Red Hook, NY, USA (2020)

\bibitem{Smolensky1990TensorPV}
Smolensky, P.: Tensor product variable binding and the representation of symbolic structures in connectionist systems. Artif. Intell.  \textbf{46},  159--216 (1990)

\bibitem{NEWELL1980}
Newell, A.: Physical symbol systems. Cognitive Science  \textbf{4}(2),  135--183 (1980)

\bibitem{PlateHolographic1995}
Plate, T.A.: Holographic reduced representations. {IEEE Transactions on Neural Networks}  \textbf{6}(3),  623--641 (1995)

\bibitem{VSA_03}
Gayler, R.W.: Vector symbolic architectures answer {J}ackendoff's challenges for cognitive neuroscience. In: {Joint International Conference on Cognitive Science (ICCS/ASCS)} (2003)

\bibitem{Palangi_AAAI2018}
Palangi, H., Smolensky, P., He, X., Deng, L.: Question-answering with grammatically-interpretable representations. In: Proceedings of the AAAI Conference on Artificial Intelligence (2018)

\bibitem{schlag2019enhancing}
Schlag, I., Smolensky, P., Fernandez, R., Jojic, N., Schmidhuber, J., Gao, J.: Enhancing the transformer with explicit relational encoding for math problem solving. arXiv preprint arXiv:1910.06611  (2019)

\bibitem{chen2020mapping}
Chen, K., Huang, Q., Palangi, H., Smolensky, P., Forbus, K., Gao, J.: Mapping natural-language problems to formal-language solutions using structured neural representations. In: International Conference on Machine Learning (ICML). pp. 1566--1575 (2020)

\bibitem{jiang2021-enrichingTPR}
Jiang, Y., Celikyilmaz, A., Smolensky, P., Soulos, P., Rao, S., Palangi, H., Fernandez, R., Smith, C., Bansal, M., Gao, J.: Enriching transformers with structured tensor-product representations for abstractive summarization. In: Proceedings of the 2021 Conference of the North American Chapter of the Association for Computational Linguistics: Human Language Technologies. pp. 4780--4793 (2021)

\bibitem{Neurocompositional}
Smolensky, P., McCoy, R.T., Fernandez, R., Goldrick, M., Gao, J.: Neurocompositional computing: From the central paradox of cognition to a new generation of ai systems. AI Magazine  \textbf{43}(3),  308--322 (2022)

\bibitem{dtm}
Soulos, P., Hu, E.J., Mccurdy, K., Chen, Y., Fernandez, R., Smolensky, P., Gao, J.: Differentiable tree operations promote compositional generalization. In: Proceedings of the 40th International Conference on Machine Learning (ICML). vol.~202, pp. 32499--32520 (2023)

\bibitem{lisp-book}
Steele, G.L.: Common LISP: The Language. Digital Press (1984)

\bibitem{switch-transformers}
Fedus, W., Zoph, B., Shazeer, N.: Switch transformers: Scaling to trillion parameter models with simple and efficient sparsity. Journal of Machine Learning Research  \textbf{23}(120),  1--39 (2022)

\bibitem{mistral-of-experts}
Jiang, A.Q., Sablayrolles, A., Roux, A., Mensch, A., Savary, B., Bamford, C., Chaplot, D.S., Casas, D.d.l., Hanna, E.B., Bressand, F., et~al.: Mixtral of experts. arXiv preprint arXiv:2401.04088  (2024)

\bibitem{sparse-moe}
Shazeer, N., Mirhoseini, A., Maziarz, K., Davis, A., Le, Q., Hinton, G., Dean, J.: Outrageously large neural networks: The sparsely-gated mixture-of-experts layer. arXiv preprint arXiv:1701.06538  (2017)

\bibitem{universal-transformers}
Dehghani, M., Gouws, S., Vinyals, O., Uszkoreit, J., Łukasz Kaiser: Universal transformers. In: International Conference on Learning Representations (ICLR) (2019)

\bibitem{dtm-beta}
Soulos, P., Conklin, H., Opper, M., Smolensky, P., Gao, J, Fernandez, R.: Compositional generalization across distributional shifts with sparse tree operations (manuscript in preparation)

\bibitem{attention-is-all-you-need}
Vaswani, A., Shazeer, N., Parmar, N., Uszkoreit, J., Jones, L., Gomez, A.N., Kaiser, L., Polosukhin, I.: Attention is all you need. In: Proceedings of the 31st International Conference on Neural Information Processing Systems (NeurIPS). p. 6000–6010 (2017)

\bibitem{adaptive-compute-time}
Graves, A.: Adaptive computation time for recurrent neural networks. arXiv preprint arXiv:1603.08983  (2016)

\bibitem{pondernet}
Banino, A., Balaguer, J., Blundell, C.: Pondernet: Learning to ponder. In: 8th ICML Workshop on Automated Machine Learning (AutoML) (2021)

\bibitem{hersche2023_learnVSA}
Hersche, M., di~Stefano, F., Sebastian, A., Hofmann, T., Rahimi, A.: Probabilistic abduction for visual abstract reasoning via learning vector-symbolic architecture formulations. 3rd Workshop on Mathematical Reasoning and AI at NeurIPS  (2023)

\end{thebibliography}

\end{document}